\documentclass[10pt,twocolumn,letterpaper]{article}

\usepackage{iccv}
\usepackage{times}
\usepackage{epsfig}
\usepackage{graphicx}
\usepackage{amsmath}
\usepackage{amssymb}

\usepackage{multirow}
\usepackage{colortbl}
\usepackage[table]{xcolor}
\usepackage[marginal]{footmisc}

\usepackage[breaklinks=true,bookmarks=false]{hyperref}

\iccvfinalcopy 

\def\iccvPaperID{2376} 

\ificcvfinal\fi
\begin{document}
\title{Focusing Attention: Towards Accurate Text Recognition in Natural Images}
\author{Zhanzhan Cheng\textsuperscript{1}\qquad\qquad
Fan Bai\textsuperscript{2}\qquad\qquad
Yunlu Xu\textsuperscript{3}\qquad\qquad  Gang Zheng\textsuperscript{1} \\
Shiliang Pu\textsuperscript{1}\qquad\qquad\qquad
Shuigeng Zhou\textsuperscript{2*}\\
\textsuperscript{1}Hikvision Research Institute, China;
\textsuperscript{2}Shanghai Key Lab of Intelligent Information\\ Processing and School of Computer Science, Fudan University, Shanghai, China;\\
\textsuperscript{3}Shanghai Jiaotong University, Shanghai, China\\
{\tt\small \{chengzhanzhan;zhenggang3;pushiliang\}@hikvision.com;}\\
{\tt\small \{baif13;sgzhou\}@fudan.edu.cn; xuyunlu1030@163.com}
}

\maketitle

\begin{abstract}
   Scene text recognition has been a hot research topic in computer vision due to its various applications. The state of the art is the attention-based encoder-decoder framework that learns the mapping between input images and output sequences in a purely data-driven way.
   However, we observe that existing attention-based methods perform poorly on complicated and/or low-quality images. One major reason is that existing methods cannot get accurate alignments between feature areas and targets for such images. We call this phenomenon ``attention drift''. 
%
   To tackle this problem, in this paper we propose the \textbf{FAN} (the abbreviation of \textbf{F}ocusing \textbf{A}ttention \textbf{N}etwork) method that employs a focusing attention mechanism to automatically draw back the drifted attention. FAN consists of two major components: an attention network (AN) that is responsible for recognizing character targets as in the existing methods, and a focusing network (FN) that is responsible for adjusting attention by evaluating whether AN pays attention properly on the target areas in the images. Furthermore, different from the existing methods, we adopt a ResNet-based network to enrich deep representations of scene text images.
%
   Extensive experiments on various benchmarks, including the IIIT5k, SVT and ICDAR datasets, show that the FAN method substantially outperforms the existing methods.

\end{abstract}

\section{Introduction}
\thispagestyle{empty}
\footnote{*Corresponding author.}
Scene text recognition has attracted much research interest of the computer vision community~\cite{jaderberg2014synthetic,neumann2012real,shi2016end}. Recognizing scene text is of great significance for scene understanding. Despite several decades of research on Optical Character Recognition (OCR), recognizing texts from natural images is still a challenging task. The state of the art employs attention-mechanism for character recognition, and achieves substantial performance improvement~\cite{lee2016recursive,shi2016robust}.

Usually, an attention-based text recognizer is designed as an encoder-decoder framework.
In the encoding stage, an image is transformed into a sequence of feature vectors by CNN/LSTM~\cite{shi2016robust}, and each feature vector corresponds to a region in the input image. In this paper, we call such regions \emph{attention regions}.
In the decoding stage, the attention network (AN) first computes alignment factors~\cite{bahdanau2014neural} by referring to the history of target characters and the encoded feature vectors for generating the synthesis vectors (also called glimpse vectors), thus achieves the alignments between the attention regions and the corresponding ground-truth-labels~\cite{bahdanau2014neural,chorowski2015attention}. Then, a recurrent neural network (RNN) is used to generate the target characters based on the glimpse vectors and the history of target characters.

\begin{figure}[!thbp]
    \begin{center}
    \includegraphics[width=0.45\textwidth]{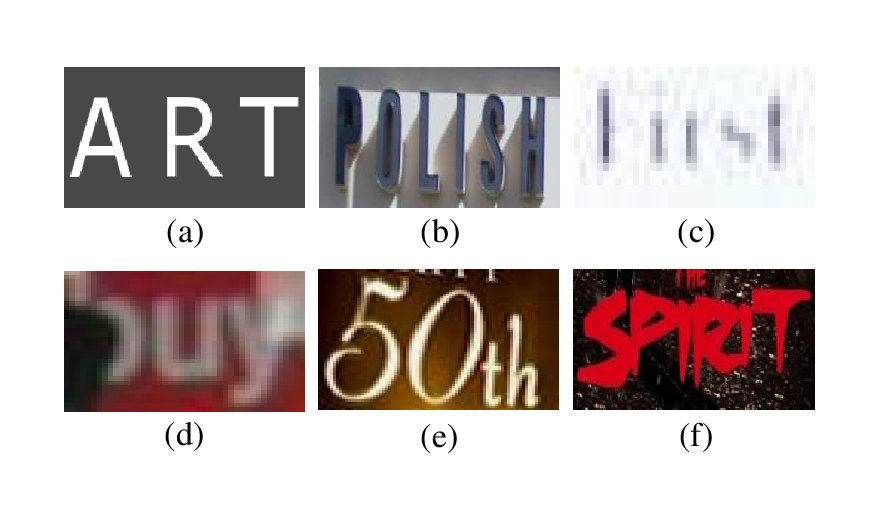}
    \end{center}
    \caption{Examples of complicated / low-quality images. Subfigures (a) - (f) represent normal, complex background, blur, incomplete, different-size and abnormal font images, respectively.
    }
    \label{fig:hard}
\end{figure}

\begin{figure}[!thbp]
    \begin{center}
    \includegraphics[width=0.45\textwidth]{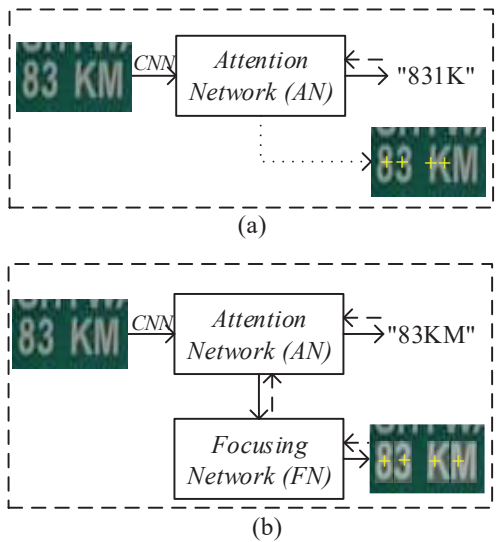}
    \end{center}
    \caption{Illustration of attention drift in the AN model and the focusing mechanism in the FAN method.
    In sub-figure (a), a real image with text ``83KM'' is taken as input, and it outputs ``831K''. The last two characters `K' and `M' are not recognized because AN's attention regions for these two characters are deviated a lot from them in the image. In sub-figure (b), with the FN component, AN's attention centers for the last two characters are rectified and positioned just on them, thus FAN outputs the correct text string ``83KM''. Here, the dotted arrows and yellow `$+$'s represent the computation of attention regions and the centers of attention regions, and the white rectangular masks in the right-bottom image indicate the ground-truth-areas of the characters.}
    \label{fig:motivation}
\end{figure}

\textbf{Motivation}.
We all know that in real scene text recognition tasks, many images are complicated (\eg distorted or overlapping characters, characters of different fonts, sizes and colors, and complex backgrounds) or low quality (due to illumination change, blur, incompleteness and noise etc.) Fig.~\ref{fig:hard} shows some examples of complicated / low-quality images. For such images,
existing attention-based methods often perform poorly. After carefully analyzing many intermediate and final results of attention-based methods on real data, we found that one major reason of poor performance is because the alignments estimated by the attention model are easily corrupted due to the complexity and/or low-quality of images. In other words, the attention model cannot accurately associate each feature vector with the corresponding target region in the input image. We call this phenomenon \emph{attention drift}. That is, the attention regions of AN deviate in some degree from the proper regions of target characters in the images.
This motivates us to develop some mechanism to focus AN's attention back on the right regions of the target characters in the input image.

Fig.~\ref{fig:motivation}(a) illustrates the attention drift phenomenon in the AN model. After the left image is input, we expect that the AN model outputs a text string ``83KM'', but actually it returns ``831K''. Note that this is not a toy example, it is a real example selected from our experiments. In practice, there are many of such examples. Obviously, the last two characters `K' and `M' are not correctly recognized. How this happens? By computing the attention regions of the four characters in the image, we can get their attention centers (the details are given in Section~\ref{sec:fn}), which are illustrated by yellow `+' in the right-bottom corner with the original image as background. We can see that the attention centers of `8' and `3' are located just over themselves, while the third attention center is on the left-half part of `K' and the fourth attention center is near the right half part of `K'. As the left-half part of `K' looks like a `1', the AN model outputs a `1'. The fourth attention region covers most part of `K', the AN model thus returns a `K'.

\textbf{Our work}.
To solve the above problem, in this paper we propose a novel method called FAN (the abbreviation of Focusing Attention Network) to accurately recognize text from natural images. Fig.~\ref{fig:motivation}(b) shows the architecture of the FAN method.
FAN is made of two major subnetworks: an attention network (AN)
that is to recognize the target characters as in existing methods, and a \emph{focusing network} (FN) that is responsible for automatically adjusting the attention of AN by first examining whether AN's attention regions are properly on the right regions of target characters in the images.
In Fig.~\ref{fig:motivation}(b), with the FN component, AN's attention regions for the last two characters are rectified, and consequently FAN outputs the correct text string ``83KM''.


Contributions of this paper are as follows:

1) We propose the concept of attention drift, which explains the poor performance of existing attention based methods on complicated and/or low-quality natural images.

2) We develop a novel method called FAN to solve the attention drift problem, where in addition to the AN component existing in most existing methods, a completely new component --- focusing network (FN) is introduced, which can focus AN's deviated attention back on the target areas.

3) We adopt a powerful ResNet-based~\cite{he2016deep} convolution neural network (CNN) to enrich deep representations of scene text images.

4) We conduct extensive experiments on several benchmarks, which demonstrate the performance superiority of our method over the existing methods. 

\section{Related work}
In recent years, there has been a lot work on scene text recognition.
For general information of text recognition, the readers can refer to Ye and Doermann's recent survey~\cite{ye2015text}.
Traditionally, there are two types of approaches: bottom-up and top-down. Early works mainly develop \emph{bottom-up} approaches: first detecting characters one by one by sliding window~\cite{wang2011end,wang2010word}, connected components~\cite{neumann2012real} or Hough voting~\cite{yao2014strokelets}, then integrating these characters into the output text.
The other approaches work in a top-down style: directly predicting the entire text from the original image without detecting the characters.
Jaderberg \etal designed a convolutional neural network (CNN) with structured output layer for unconstrained recognition~\cite{jaderberg2014deep}.
They also conducted a 90k-class classification task with a CNN, in which each class represents an English word~\cite{jaderberg2016reading} .
Recent works solve this problem as a sequence recognition problem, where images and texts are separately encoded as patch sequences and character sequences, respectively.
Sutskever \etal\cite{sutskever2014sequence} extracted sequences of HOG features to represent images, and generated the character sequence with the recurrent neural network (RNN).
Shi \etal\cite{shi2016end} proposed an end-to-end neural network that combines CNN and RNN. They also developed an attention-based spatial transformer network (STN) for rectifying text distortion, which helps recognize curved scene texts~\cite{shi2016robust}.

Different from the existing approaches, in this paper we first extract deeper representations of images by using a ResNet-based CNN. To the best of our knowledge, this may be the first work of scene text recognition that uses a ResNet-based CNN. Then, we feed the sequence of features to an AN for generating alignment factors and \emph{glimpse vectors}. Meanwhile, we employ a FN to evaluate whether the \emph{glimpse vectors} are reasonable and provide a feedback to help AN generate more reasonable \emph{glimpse vectors} so that the AN can pay attention properly on the right regions of target characters in the processed image.

Though \emph{attention drift} has been observed in attention training of speech recognition~\cite{kim2016joint}, where the authors proposed an MTL framework that combines CTC and AN to handle this issue, our paper is the first work that formally puts forward the concept of \emph{attention drift}. Furthermore, we design a focus-mechanism to solve this problem. 
It is worth of noting that we have tried to use CTC and AN to solve the attention drift problem in scene text recognition, unfortunately our extensive experiments showed that this idea does not work well, so we discarded it.
\section{The FAN Method}
As shown in Fig.~\ref{fig:motivation}(b), FAN has two major modules: AN and FN.
In the AN component, alignment factors~\cite{bahdanau2014neural} between target labels and
features are generated. Each alignment factor corresponds to an attention region in the input image. Bad alignments (\ie deviated or unfocused attention regions) result in poor recognition results. 
The FN component first locates the attention region for each target label, then conducts dense prediction over the attention region with the corresponding \emph{glimpse vector}. In such a way, FN can judge whether the glimpse vector is reasonable.
In summary, FN generates the dense outputs over the attention regions in the input image based on the \emph{glimpse vectors} provided by AN, and AN in turn updates the \emph{glimpse vectors} based on FN's feedback.


\subsection{Attention Network (AN)}\label{sec:an}
An attention-based decoder is a recurrent neural network~(RNN) that directly generates the target sequence $(y_1, ..., y_M)$ from an input image $\mathcal{I}$. 
In practice, image $\mathcal{I}$ is often encoded as a sequence of feature vectors by CNN-LSTM. Formally, $Encoder(\mathcal{I})$=$(h_1, ..., h_T)$. Bahdanau \etal\cite{bahdanau2014neural} first proposed the architecture of attention-based decoder.
At the $t\text{-}th$ step, the decoder generates an output $y_t$

\begin{equation}
y_t = Generate(s_t, g_t),
\label{eq:py}
\end{equation}
where $s_t$ is an RNN hidden state at time $t$, computed by

\begin{equation}
s_t = RNN(y_{t-1}, g_t, s_{t-1}),
\label{eq:rnn}
\end{equation}
and $g_t$ is the weighted sum of sequential feature vectors $(h_1, ..., h_T)$, that is,

\begin{equation}
g_t = \sum_{j=1}^T \alpha_{t,j}h_j,
\end{equation}
where $\alpha_t \in \mathbb{R}^T$ is a vector of \textit{attention weights}, also called \textit{alignment factors}. 
$\alpha_t$ is often evaluated by scoring each element in $(h_1, ..., h_T)$ separately and then normalizing the scores as follows:
\begin{equation}
e_{t,j} = v^{T} \tanh(Ws_{t-1} + Vh_j + b),
\end{equation}

\begin{equation}
\alpha_{t,j} = \frac{exp(e_{t,j})}{\sum_{j=1}^T exp(e_{t,j})}.
\end{equation}
Above, $v$, $W$, $V$ and $b$ are all trainable parameters.

Here, the $Generate$ function in Eq.~(\ref{eq:py}) and the $RNN$ function in Eq.~(\ref{eq:rnn}) represent a feed-forward network and a LSTM recurrent network, respectively.
Besides, the decoder needs to generate sequences of variable lengths. Following \cite{sutskever2014sequence}, a special end-of-sentence (EOS) token is added to the target set, so that the decoder completes the generation of characters when EOS is emitted. The loss function of the attention model is as follows:
\begin{equation}\label{eq:loss_attention}
\mathcal{L}_{Att} = -\sum_{t}ln P(\hat{y}_t | \mathcal{I}, \theta),
\end{equation}
where $\hat{y}_t$ is the ground truth of the $t\text{-}th$ character and $\theta$ is a vector that combines all the network parameters.

\begin{figure*}[!htbp]
    \begin{center}
    \includegraphics[width=0.75\textwidth]{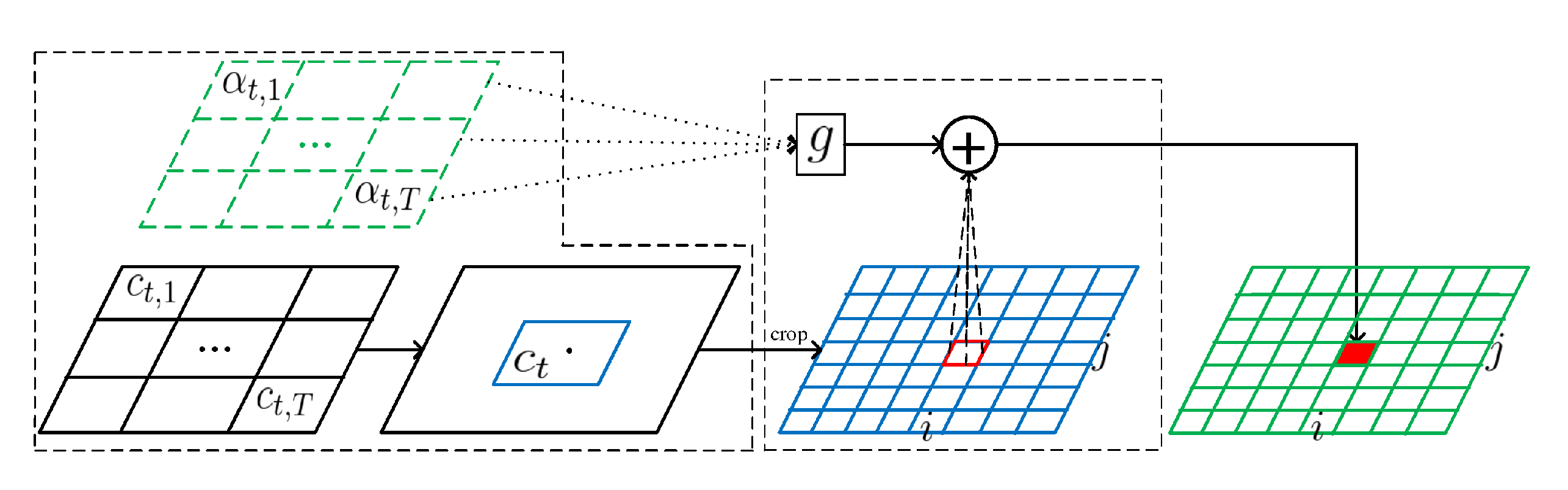}
    \end{center}
    \caption{The mechanism of attention focusing in FAN. Here, $\alpha$, $c$, $g$ and $+$ represent alignment factors,  center of each feature in the input image, \emph{glimpse vector} and focusing operation, respectively. The blue grid and the green grid indicate the cropped features and predicted results over each pixel, respectively.
    In order to predict the $t\text{-}th$ target,
    we first evaluate the center position $c_{t,j}$ for each feature vector $h_j$ obtained by \emph{CNN-LSTM},
    and compute weighted sum of all centers to get a weighted position $c_{t}$, then crop a patch of features from the input image or a convolution output and do the focusing operation over the attention region.
    }
    \label{fig:focus}
\end{figure*}

In scene text recognition, the AN model has two major drawbacks: 1)
This model is easily impacted by complicated / low-quality scene data,
and generates imprecise alignment factors because the model has no alignment constraint on the integration of \emph{glimpse vectors}, which may result in mismatch between attention regions and ground-truth regions. This is the so-called attention drift problem mentioned above. 2) It is hard to train such a model on huge scene text data, such as the 800-million synthetic data released by Gupta \etal\cite{Gupta16}. In this paper, our major goal is to tackle the attention drift problem. We try to constrain AN's attention just on each target character by introducing the focusing network, which is detailed in the following section.

\subsection{Focusing Network (FN)}\label{sec:fn}
In the attention model, each feature vector is mapped to an area of the input image, which can be used to localize the target character based on the convolution strategy.
However, the computed attention regions of targets are usual inaccurate, especially for complicated and/or low-quality images.
In order to handle the \textit{attention drift} problem, we introduce a focusing network with
a focusing-mechanism to rectify the drifted attention. The focusing-mechanism is illustrated in Fig.~\ref{fig:focus}. 
It works in two major steps: 1) computing the attention center of each predicted label; 2) focusing attention on target regions by generating the probability distributions on the attention regions.

{\bf{Computing attention center}}:
In convolution/pooling operation, we define the input as $N \times D_i \times H_i \times W_i$, and the output as $N \times D_o \times H_o \times W_o$, where $N$, $D$, $H$ and $W$ indicate the batch size, the number of channels, the height and width of feature maps, respectively. With the convolution strategy: $kernel$, $stride$ and $pad$, we have $H_o$ = $(H_i + 2 \times pad_H - kernel_H) / stride_H + 1$ and $W_o$ = $(W_i + 2 \times pad_W - kernel_W) / stride_W + 1$. Therefore, for position $(x, y)$ in layer $L$, we compute its receptive field in layer $L-1$ as the bounding box coordinates $r = (x_{min}, x_{max}, y_{min}, y_{max})$ as follows:
\begin{equation}
\begin{split}
&   x_{min} = (x - 1) \times stride_W + 1 - pad_W, \\
&   x_{max} = (x - 1) \times stride_W - pad_W + kernel_W, \\
&   y_{min} = (y - 1) \times stride_H + 1 - pad_H, \\
&   y_{max} = (y - 1) \times stride_H - pad_H + kernel_H.
\end{split}
\label{eq:up}
\end{equation}

At the $t$-th step, we compute the receptive field of $h_j$ in the input image by recursively performing the computation of Eq.~ (\ref{eq:up}), and select the center of the receptive field as the attention center:
\begin{equation}
c_{t,j} = location(j)
\end{equation}
where $j$ refers the index of $h_j$, and $location$ denotes the function of evaluating the center of a receptive field.
Therefore, the attention center of target $y_t$ in the input image is evaluated as follows:
\begin{equation}
c_t = \sum_{j=1}^T \alpha_{t,j} c_{t,j}.
\end{equation}

{\bf{Focusing attention on target regions}}:
With the computed attention center of target $y_t$, we crop a patch of feature maps of size $\mathcal{P}(\mathcal{P}_H, \mathcal{P}_W)$ from the input image or a convolution output as follows:
\begin{equation}
\mathcal{F}_t = Crop(\mathcal{F}, c_t, \mathcal{P}_H, \mathcal{P}_W)
\end{equation}
where $\mathcal{F}$ is the image/convolution feature maps, $\mathcal{P}$ is set to the max-size of ground-truth regions in the input image.

With the cropped feature maps, we compute the energy distribution over the attention region as follows:
\begin{equation}
e_t^{(i,j)} = \tanh(Rg_t + S\mathcal{F}_t^{(i,j)} + b)
\end{equation}
Above, $R$ and $S$ are trainable parameters, and $(i,j)$ refers to the $(i \times \mathcal{P}_W + j)$-th feature vector.
Then, the probability distribution over the selected region is computed as
\begin{equation}
P_t^{(i,j, k)} = \frac{exp(e_t^{(i,j,k)})}{\sum_{k^\prime}^K exp(e_t^{(i,j,{k^\prime})})},
\end{equation}
where $K$ is the number of label classes.

Then, we define the \emph{focusing loss function} as
\begin{equation}
\mathcal{L}_{focus} = - \sum_t^M \sum_i^{\mathcal{P}_W} \sum_j^{\mathcal{P}_H} log P(\hat{y}_{t}^{(i,j)} | \mathcal{I}, \omega)
\label{eq:loss_focus}
\end{equation}
where $\hat{y}_{t}^{(i,j)} $ is the ground-truth pixel label and $\omega$ is a
vector that combines all the FN parameters.  The loss is added only for the subset of images with character annotations.
\subsection{FAN Training}\label{sec:training}
\begin{figure}[!thbp]
    \begin{center}
    \includegraphics[width=0.45\textwidth]{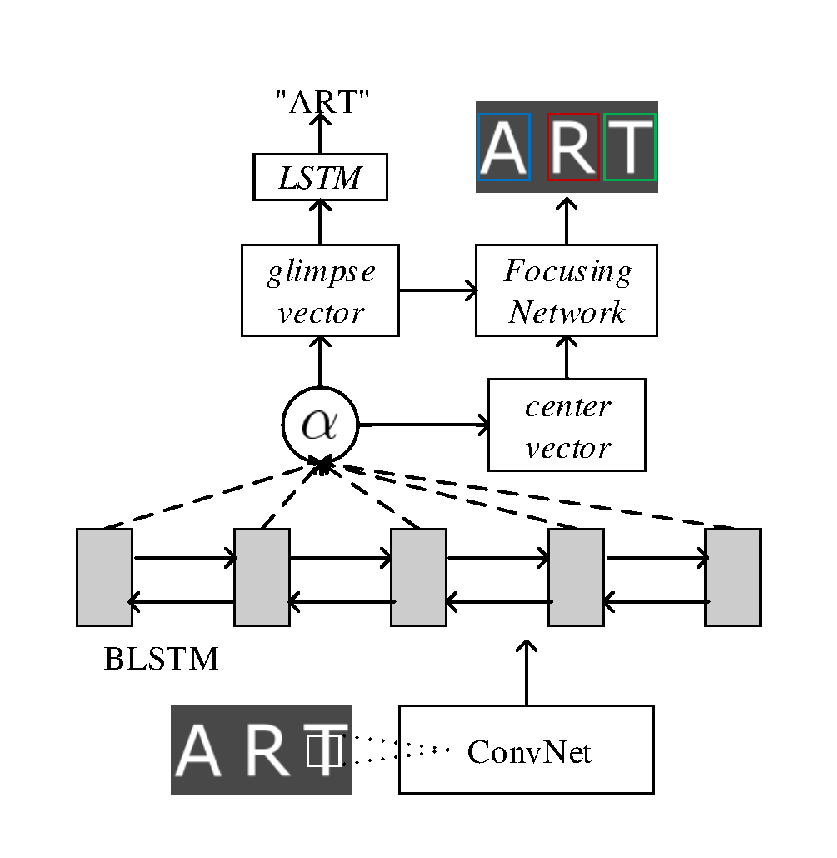}
    \end{center}
    \caption{The FAN network architecture. Here, the \emph{CNN-BLSTM} encoder transforms an input image $\mathcal{I}$ into high level sequence of features, the RNN decoder generates each target character, and FN focuses the attention of AN on the right target character regions in the input images. FN and AN are trained simultaneously.}
    \label{fig:attention_focus}
\end{figure}

We combine a ResNet-based feature extractor, AN and FN into one network, as shown in Fig.~\ref{fig:attention_focus}.
The details are given in Section~\ref{sec:implementation}. AN uses the extracted features to generate alignment factors and glimpse vectors, with which FN focuses the attention of AN on the proper target character regions in the images.

The objective function is constructed by considering both target-generation and attention-focusing as follows:
\begin{equation}\label{eq:training-objective}
\mathcal{L} = (1 - \lambda) \mathcal{L}_{Att} + \lambda\mathcal{L}_{focus},
\end{equation}
with a tunable parameter $\lambda$ ($0 \leqslant \lambda < 1$), which trades off the impact of AN and FN.
The network is trained by standard back-propagation~\cite{rumelhart1988learning}.

\subsection{Decoding}
The attention-based decoder is to generate the output sequence of characters from the implicitly learned character-level probability statistics. In the process of unconstrained text recognition (lexicon-free), we straightforwardly select the most probable character. While in constrained text recognition, according to the lexicons of different sizes (denoted by ``50'', ``1k'' and ``full''), we calculate the conditional probability distributions for all lexicon words, and take the one with the highest probability as output result.
\begin{table}[!htbp]
\begin{center}
    \begin{tabular}{c|c|c}
    \hline
    layer name  &32 layers      & output size \cr\hline


    \multirow{2}{*}{conv1\_x} &
      $3\times3, 1\times1, 1\times1, 32$ & \multirow{2}{*}{$32\times256$} \cr\cline{2-2}
    & $3\times3, 1\times1, 1\times1, 64$ &   \cr\hline

    &pool:$2\times2, 2\times2,0\times0$ & \cr\cline{2-2}
    & \cellcolor{gray!10}$\begin{bmatrix}3\times3,128\\3\times3,128\end{bmatrix}\quad$$\times$1&  \cr\cline{2-2}
    \multirow{-4}{*}{conv2\_x} & $3\times3, 1\times1, 1\times1, 128$ & \multirow{-4}{*}{$16\times128$}   \cr\hline

    \multirow{4}{*}{conv3\_x} &
    pool:$2\times2, 2\times2,0\times0$ & \multirow{4}{*}{$8\times64$} \cr\cline{2-2}
    & \cellcolor{gray!10}$\begin{bmatrix}3\times3,256\\3\times3,256\end{bmatrix}\quad$$\times$2&  \cr\cline{2-2}
    & $3\times3, 1\times1, 1\times1,256$ &  \cr\hline

    \multirow{4}{*}{conv4\_x} &
    pool:$2\times2, 1\times2,1\times0$ & \multirow{4}{*}{$4\times65$} \cr\cline{2-2}
    & \cellcolor{gray!10}$\begin{bmatrix}3\times3,512\\3\times3,512\end{bmatrix}\quad$$\times$5&  \cr\cline{2-2}
    & $3\times3, 1\times1, 1\times1, 512$ &  \cr\hline

    & \cellcolor{gray!10}$\begin{bmatrix}3\times3,512\\3\times3,512\end{bmatrix}\quad$$\times$3&  \cr\cline{2-2}
    & $2\times2, 1\times2, 1\times0, 512$ &  \cr\cline{2-2}
    \multirow{-4}{*}{conv5\_x} & $2\times2, 1\times1, 0\times0, 512$ & \multirow{-4}{*}{$1\times65$} \cr\hline
    \end{tabular}
  \end{center}
  \caption{A ResNet-based CNN architecture for robust text feature extraction. Building blocks are shown in brackets, and ResNet blocks are highlighted with gray background.}
  \label{tab:architectures}
\end{table}
\section{Performance Evaluation}\label{sec:experiments}
We conduct extensive experiments to validate the proposed FAN method on a number of general recognition benchmarks commonly used in the literature. For comprehensive performance comparison, FAN is compared with 18 existing methods and the AN model implemented with a ResNet-based encoder, which is taken as the baseline method. To further verify the effectiveness of our attention focusing mechanism, we compare FAN with AN in the context that both are based on the image-encoder released by Shi \etal\cite{shi2016robust}. Finally, we also investigate the effect of two major parameters on the performance of FAN.

\begin{table*}[!htbp]
  \begin{center}
    \begin{tabular}{|l||c|c|c||c|c||c|c|c||c||c|}
    \hline
    \multirow{2}{*}{\textbf{Method}} &
    \multicolumn{3}{c||}{\textbf{IIIT5k}} & \multicolumn{2}{c||}{\textbf{SVT}} & \multicolumn{3}{c||}{\textbf{IC03}} & \textbf{IC13} & \textbf{IC15} \cr\cline{2-11}
      & \textbf{50} & \textbf{1k} & \textbf{None} & \textbf{50} & \textbf{None} & \textbf{50} & \textbf{Full} & \textbf{None} & \textbf{None} & \textbf{None} \cr\hline
      ABBYY~\cite{wang2011end} & 24.3          & $-$ & $-$ & 35.0 & $-$ & 56.0 & 55.0 & $-$ & $-$  & $-$\cr
      Wang \etal\cite{wang2011end}           & $-$  & $-$  & $-$ & 57.0 & $-$ & 76.0 & 62.0 & $-$ & $-$  & $-$\cr
      Mishra \etal\cite{graves2013speech}         & 64.1 & 57.5 & $-$ & 73.2 & $-$ & 81.8 & 67.8 & $-$ & $-$  & $-$\cr
      Wang \etal\cite{wang2012end}           &$-$  & $-$  & $-$ & 70.0 & $-$ & 90.0 & 84.0 & $-$ & $-$  & $-$\cr
      Goel \etal\cite{goel2013whole}           &$-$  & $-$  & $-$ & 77.3& $-$ & 89.7& $-$  & $-$ & $-$  & $-$\cr
      Bissacco \etal\cite{bissacco2013photoocr}       &$-$  & $-$  & $-$ & 90.4& 78.0& $-$ & $-$ & $-$ & 87.6 & $-$\cr
      Alsharif and Pineau~\cite{alsharif2013end}   &$-$  & $-$  & $-$ & 74.3& $-$ & 93.1 & 88.6 & $-$ & $-$  & $-$\cr
      Almaz{\'a}n \etal\cite{almazan2014word}         &91.2  & 82.1 & $-$ & 89.2 & $-$ & $-$ & $-$ & $-$ & $-$  & $-$\cr
      Yao \etal\cite{yao2014strokelets}            &80.2 & 69.3 & $-$ & 75.9& $-$ & 88.5 & 80.3 & $-$ & $-$  & $-$\cr
  Rodr{\'i}guez-Serrano \etal\cite{rodriguez2015label}  &76.1 & 57.4 & $-$ & 70.0& $-$ & $-$ & $-$ &  $-$ & $-$  & $-$\cr
      Jaderberg \etal\cite{jaderberg2014deep}      &$-$  & $-$  & $-$ & 86.1& $-$ & 96.2& 91.5 & $-$ & $-$  & $-$\cr
      Su and Lu~\cite{su2014accurate}             &$-$  & $-$  & $-$ & 83.0& $-$ & 92.0& 82.0 & $-$ & $-$  & $-$\cr
      Gordo~\cite{gordo2015supervised}                 &93.3 & 86.6 & $-$ & 91.8& $-$ & $-$ & $-$ & $-$ & $-$  & $-$\cr
      Jaderberg \etal\cite{jaderberg2016reading}      &97.1 & 92.7 & $-$ & 95.4& 80.7& 98.7& \textbf{98.6}& 93.1& 90.8 & $-$\cr
      Jaderberg \etal\cite{jaderberg2014deep}      &95.5 & 89.6 & $-$ & 93.2& 71.7& 97.8& 97.0& 89.6& 81.8 & $-$\cr
      Shi \etal\cite{shi2016end}            &97.6 & 94.4 & 78.2& 96.4& 80.8& 98.7& 97.6& 89.4& 86.7 & $-$\cr
      Shi \etal\cite{shi2016robust}            &96.2 & 93.8 & 81.9& 95.5& 81.9& 98.3& 96.2&  90.1& 88.6 & $-$\cr
      Baidu IDL.~\cite{baidu2016joint}            &$-$ & $-$ & $-$& $-$& $-$& $-$& $-$& $-$& $89.9$ & \textbf{77.0}\cr\hline
      Baseline        &98.9 & 96.8 & 83.7& 95.7& 82.2& 98.5& 96.7&  91.5& 89.4 & 63.3\cr
      FAN &\textbf{99.3} & \textbf{97.5} & \textbf{87.4}& \textbf{97.1}& \textbf{85.9}& \textbf{99.2}& 97.3& \textbf{94.2}& \textbf{93.3} & 70.6 \cr\hline
    \end{tabular}
    \end{center}
  \caption{Recognition accuracies on general benchmarks. ``50'' and ``1k'' are lexicon sizes, ``Full'' indicates the combined lexicon of all images in the benchmarks, and ``None'' means lexicon-free.}
  \label{tab:results}
\end{table*}

\subsection{Datasets}
{\bf{IIIT 5K-Words}} \cite{mishra2012scene} (IIIT5K) is collected from the Internet, containing 3000 cropped word images in its test set. Each image specifies a 50-word lexicon and a 1k-word lexicon, both of which contain the ground truth words as well as other randomly picked words.

{\bf{Street View Text}} \cite{wang2011end} (SVT) is collected from the Google Street View, consists of 647 word images in its test set. Many images are severely corrupted by noise and blur, or have very low resolutions. Each image is associated with a 50-word lexicon.

{\bf{ICDAR 2003}} \cite{lucas2003icdar} (IC03) contains 251 scene images, labeled with text bounding boxes. Each image is associated with a 50-word lexicon defined by Wang \etal \cite{wang2011end}. For fair comparison, we discard images that contain non-alphanumeric characters or have less than three characters, following \cite{wang2011end}. The resulting dataset contains 867 cropped images. The lexicons include the 50-word lexicons and the full lexicon which combines all lexicon words.

{\bf{ICDAR 2013}} \cite{karatzas2013icdar} (IC13) is the successor of IC03, from which most of its data are inherited. It contains 1015 cropped text images. No lexicon is associated.

{\bf{ICDAR 2015}} \cite{karatzas2015icdar} (IC15) contains 2077 cropped images. For fair comparison, we discard the images that contain non-alphanumeric characters or irregular (arbitrary oriented, perspective or curved) texts, which results in 1811 images. No lexicon is associated.

\subsection{Implementation Details}\label{sec:implementation}
{\bf{Network:}} For the encoder, we construct a 32-layer ResNet~\cite{he2016deep}-based CNN as described in Tab.~\ref{tab:architectures} for extracting deeper represented text features.
All the ResNet blocks with gray background in Tab.~\ref{tab:architectures} have the following format: $\{[kernel~size, number~of~channels] \times\}$, each of which has $\{stride, pad\}=\{0, 0\}$.
The other convolution layers have the following format: $\{kernel_W \times kernel_H, stride_W \times stride_H, pad_W \times pad_H, channels\}$; And the maxpooling layers have the following format: $pool:\{kernel_W \times kerne_H, stride_W \times stride_H, pad_W \times pad_H\}$. The subscript $H$ and $W$ respectively represent the height and width of feature maps.
Then, the extracted sequence of features from CNN are fed into a BLSTM~(256 hidden units) network.
For the character generation task, the attention is designed with a LSTM~(256 memory blocks) and 37 output units (26 letters, 10 digits, and 1 EOS symbol).
For FAN, we crop feature maps from the input image and set $\lambda = 0.01$.
We will further discuss the tuning of parameter $\lambda$ in Section~\ref{sec:parameter-tuning}.

\begin{table*}[!htbp]
  \begin{center}
    \begin{tabular}{|l||c|c|c||c|c||c|c|c||c||c|}
    \hline
    \multirow{2}{*}{\textbf{Method}} &
    \multicolumn{3}{c||}{\textbf{IIIT5k}} & \multicolumn{2}{c||}{\textbf{SVT}} & \multicolumn{3}{c||}{\textbf{IC03}} & \textbf{IC13} & \textbf{IC15} \cr\cline{2-11}
      & \textbf{50} & \textbf{1k} & \textbf{None} & \textbf{50} & \textbf{None} & \textbf{50} & \textbf{Full} & \textbf{None} & \textbf{None} & \textbf{None} \cr\hline
      Baseline        &31.7 & 78.0 & 188.9 & 22.9 & 47.3 & 11.2 & 18.4 &  28.5 & 58.1 & 306.8\cr
      FAN                  &\textbf{22.0} & \textbf{64.5} & \textbf{145.2}& \textbf{16.3}& \textbf{36.7}& \textbf{6.0}& \textbf{16.2}& \textbf{21.3}& \textbf{39.8} & \textbf{251.9} \cr\hline
    \end{tabular}
  \end{center}
  \caption{The total normalized edit distance results on general benchmarks. ``50'' and ``1k'' are lexicon sizes, ``Full'' indicates the combined lexicon of all images in the dataset, and ``None'' means lexicon-free.}
  \label{tab:edit}
\end{table*}

{\bf{Model Training:}}
With the ADADELTA \cite{adadelta} optimization method, we train our model on
8-million synthetic data without being pixel-wise labeled released by Jaderberg \etal\cite{jaderberg2014synthetic} and
4-million synthetic pixel-wise labeled word instances (excluding the images that contain non-alphanumeric characters) cropped from 80-thousand images released by Gupta \etal\cite{Gupta16}. That is, about 30\% instances have pixel labels.
We set the batch size to 32 and scale images to $256\times32$ in both training and testing.
Our model processes about 90 samples per second, and converges in 5 days after about 3 epochs over the training set.

{\bf{Implementation and Running Environment}}
Our method is implemented under the CAFFE framework \cite{jia2014caffe}.
We use the CUDA backend extensively in our implementation, so that most modules in our model are GPU-accelerated.
Our experiments are carried out on a workstation with one Intel Xeon(R) E5-2650 2.30GHz CPU, an NVIDIA Tesla M40 GPU, and 128GB RAM.

\subsection{Performance on General Recognition Datasets}
Tab.~\ref{tab:results} shows the performance results of our method and 18 existing methods as well as a \emph{baseline method} that combines an AN and a ResNet-based feature extractor. 
The only difference between the FAN method and the baseline is that FAN has an FN component. So comparing the results of the baseline and FAN, we can infer the performance contribution of the FN component to the FAN method.

From Tab.~\ref{tab:results}, we can see that FAN achieves better performance than the baseline in all cases, and substantially outperforms the 18 existing methods on almost all benchmarks, but IC03 with `Full' lexicon and IC15. This demonstrates the effectiveness and superiority of the FAN method. Moreover, for both constrained and unconstrained cases, the baseline method significantly outperforms all the existing methods on IIIT5k, SVT, and performs comparably to the best existing method on IC03 and IC13, but falls behind \cite{baidu2016joint} released on the ICDAR15 competition. In fact, our model don't conduct fine-tune on the IC15 train set.
Thus we can conclude that the ResNet-based network indeed extracts more robust features for scene text recognition and greatly boosts the recognition performance.



In the ICDAR competition,
the sum of normalized edit distance (NED)~\cite{karatzas2015icdar} of each image is used as the final indicator for method ranking.
We also evaluate the total normalized edit distance (NED) on all benchmarks for the baseline and FAN, and the results are shown in Tab.~\ref{tab:edit}. Here, NED is simply defined as ${edit\_distance(pred, gt)}/{|gt|}$, where $pred$ and $gt$ represent the prediction result and ground truth respectively. Comparing to the baseline method, we can see that FAN significantly improves the NED measure in both constrained and unconstrained cases.
\begin{figure}[!thbp]
    \begin{center}
    \includegraphics[width=0.35\textwidth]{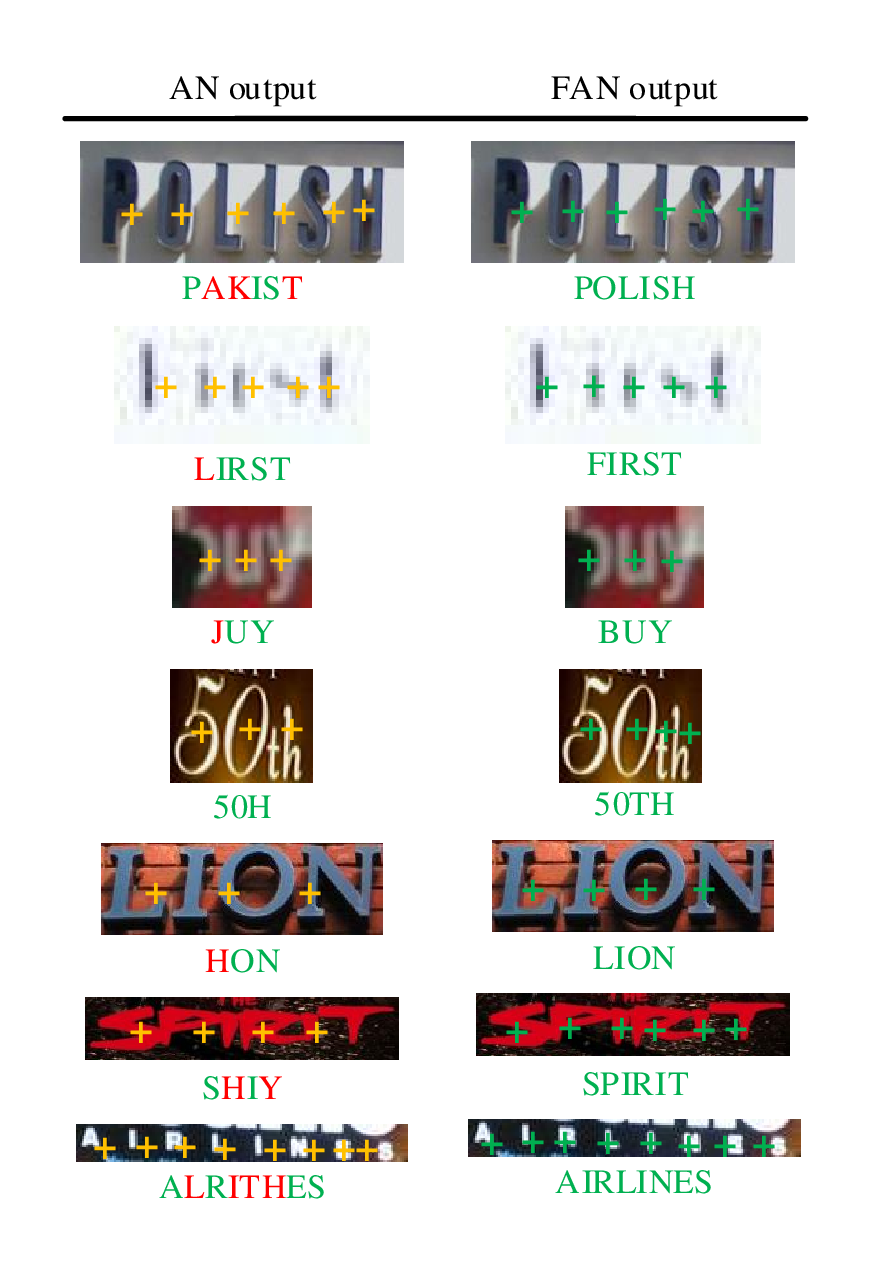}
    \end{center}
    \caption{Some real images processed by AN and FAN. The left and right columns are the output results of AN and FAN respectively. The predicted text string for each input image is shown just below the image. The attention centers are marked as `$+$' of yellow (for AN) and green (for FAN). Green/red characters indicate correctly/incorrectly recognized characters. 
    }
    \label{fig:visual}
\end{figure}

Fig.~\ref{fig:visual} shows some real images processed by AN and FAN.
Comparing with the outputs of AN, FAN obviously rectify the attention drift problem and correctly recognize more characters in the images.

\begin{table}[!htbp]
  \begin{center}
    \begin{tabular}{|l|c|c|c|c|c|}
    \hline
    \textbf{Method} & \textbf{IIIT5k} & \textbf{SVT} & \textbf{IC03} & \textbf{IC13} & \textbf{IC15} \cr\hline
      AN        & 248.0 & 65.9 & 43.1 & 85.2 & 348.2 \cr
      FAN             &\textbf{213.4} & \textbf{56.0} & \textbf{31.5}& \textbf{72.1}& \textbf{323.6}\cr\hline
    \end{tabular}
  \end{center}
  \caption{Total NED results of AN and FAN on unconstrained benchmarks with the image-encoder released by Shi \etal\cite{shi2016robust}.}
  \label{tab:encoder-ned}
\end{table}

\begin{table}[!htbp]
  \begin{center}
    \begin{tabular}{|l|c|c|c|c|c|}
    \hline
    \textbf{Method} & \textbf{IIIT5k} & \textbf{SVT} & \textbf{IC03} & \textbf{IC13} & \textbf{IC15} \cr\hline
      AN        & 79.5 & 75.6 & 88.2 & 85.8 & 58.2 \cr
      FAN             &\textbf{81.7} & \textbf{78.7} & \textbf{90.3}& \textbf{87.8}& \textbf{61.0}\cr\hline
    \end{tabular}
  \end{center}
  \caption{Accuracy results of AN and FAN on unconstrained benchmarks with the image-encoder released by Shi \etal\cite{shi2016robust}.}
  \label{tab:encoder-acc}
\end{table}

As we know, FAN consists of three major components: AN, FN and a ResNet-based encoder.
To further evaluate the performance contribution of FN to FAN, here we compare the performance of FAN and AN, which are all based on the image-encoder released by Shi \etal\cite{shi2016robust}, instead of the ResNet-based encoder, over the five unconstrained datasets (IIIT5K, SVT, IC03, IC13 and IC15). That is, both FAN and AN use the image-encoder released by Shi \etal\cite{shi2016robust} to do text recognition. The results are presented in Tab.~\ref{tab:encoder-ned} and Tab~\ref{tab:encoder-acc}, which show that FAN still significantly outperforms AN in terms of accuracy and total NED. This demonstrates again the outstanding performance of our attention focusing mechanism even when the ResNet-based encoder is not used.
\subsection{The Effect of Parameter $\lambda$}\label{sec:parameter-tuning}
We need to tune the super-parameter $\lambda$ in our training objective function Eq.~(\ref{eq:training-objective}) for better performance of the network. $\lambda$ trades off the impact of AN and FN. Here, we explore how $\lambda$ impacts FAN's performance. 

\begin{table}[!htbp]
  \begin{center}
    \begin{tabular}{|c|c|c|c|c|c|}
    \hline
    $\lambda$ & IIIT5k & SVT & IC03 & IC13 & IC15 \cr\hline
    \hline
    $0.001$  & 153.2 & 40.5 & 23.3 & 44.6 & 262.9 \cr\hline
    $0.005$  & 147.5 & 37.6 & 22.8 & 42.2 & 255.3 \cr\hline
    $0.01$  & \textbf{145.2} & \textbf{36.7} & \textbf{21.3} & \textbf{39.8} & \textbf{251.9} \cr\hline
    $0.05$  & 148.4 & 38.1 & 23.1 & 45.2 & 252.5 \cr\hline
    $0.1$  & 150.0 & 38.4 & 23.3 & 49.8 & 253.8 \cr\hline
    \end{tabular}
  \end{center}
  \caption{The total NED results on unconstrained benchmarks with different $\lambda$ values.}
  \label{llambda}
\end{table}

\begin{table}[!htbp]
  \begin{center}
    \begin{tabular}{|c|c|c|c|c|c|}
    \hline
    $\lambda$ & IIIT5k & SVT & IC03 & IC13 & IC15 \cr\hline
    \hline
    $0.001$  & 86.6 & 85.8 & 93.2 & 92.7 & 69.5 \cr\hline
    $0.005$  & 87.2 & 85.6 & 93.5 & 92.8 & 70.2 \cr\hline
    $0.01$  & \textbf{87.4} & 85.9 & \textbf{94.2} & \textbf{93.3} & 70.6 \cr\hline
    $0.05$  & 87.0 & \textbf{86.2} & 94.1 & 92.5 & 70.6 \cr\hline
    $0.1$  & 86.9 & 86.2 & 93.9 & 92.1 & \textbf{70.7} \cr\hline
    \end{tabular}
  \end{center}
  \caption{Accuracy results on unconstrained benchmarks with different $\lambda$ values.}
  \label{llambda1}
\end{table}

Intuitively, $\lambda$ should be from 0 to 1. A larger value means the FN component plays a more important role in the text recognition task, and vice versa. $\lambda = 0$ means that FN does not work in the recognition process, \ie, the focusing mechanism is not employed. On the contrary, $\lambda = 1$ means that AN does not work, which is unacceptable. 
In our experiments, we vary $\lambda$ from 0 to 0.1, and the results on the unconstrained recognition benchmarks are given in Tab.~\ref{llambda} and Tab.~\ref{llambda1}. We can see that FAN performs stably and achieves relatively higher performance with $\lambda = 0.01$.

\subsection{The Effect of Pixel Labeling}
Pixel labeling for characters in the training images certainly benefits recognition performance, but also consumes large amount of resource (time and money). Therefore, here we examine how the amount of pixel labeling impacts the recognition performance. We expect that FAN trained with a small number of pixel-labeled (or positioned) samples can still achieve a high performance.

\begin{table}[!htbp]
  \begin{center}
    \begin{tabular}{|c|c|c|c|c|c|}
    \hline
    $Ratio$ & IIIT5k & SVT & IC03 & IC13 & IC15 \cr\hline
    \hline
    $0\%$  & 188.9 & 47.3 & 28.5 & 58.1 & 306.8 \cr\hline
    $1\%$  & 155.3 & 41.7 & 22.6 & 46.3 & 258.3 \cr\hline
    $5\%$  & 152.5 & 41.8 & 22.3 & 52.7 & 258.3 \cr\hline
    $10\%$  & 152.9 & 39.7 & 22.1 & 43.5 & 264.9 \cr\hline
    $30\%$  & \textbf{145.2} & \textbf{36.7} & \textbf{21.3} & \textbf{39.8} & \textbf{251.9} \cr\hline
    \end{tabular}
  \end{center}
  \caption{The total NED results  on unconstrained benchmarks for different numbers of positioned-samples.}
  \label{tab:free}
\end{table}

\begin{table}[!htbp]
  \begin{center}
    \begin{tabular}{|c|c|c|c|c|c|}
    \hline
    $Ratio$ & IIIT5k & SVT & IC03 & IC13 & IC15 \cr\hline
    \hline
    $0\%$  & 83.7 & 82.2 & 91.5 & 89.4 & 63.3 \cr\hline
    $1\%$  & 86.2 & 84.4 & 93.8 & 92.7 & 70.0 \cr\hline
    $5\%$  & 86.3 & 84.7 & 93.8 & 92.6 & 69.4 \cr\hline
    $10\%$  & 86.3 & 84.5 & 94.0 & 93.1 & 70.0 \cr\hline
    $30\%$  & \textbf{87.4} & \textbf{85.9} & \textbf{94.2} & \textbf{93.3} & \textbf{70.6} \cr\hline
    \end{tabular}
  \end{center}
  \caption{Accuracy results on unconstrained benchmarks for different numbers of positioned-samples.}
  \label{tab:free1}
\end{table}

Tab.~\ref{tab:free} and Tab.~\ref{tab:free1} show the results of NED and accuracy on unconstrained benchmarks by increasing the ratio of pixel-labeled samples from 0 to 30\%. We can see that even with only a small fraction 
of pixel-labeled samples in the training set, our method can still guide the model to achieve satisfactory performance.

\section{Conclusion}
In this work, we put forward the \emph{attention drift} concept to explain the poor performance of existing AN based methods of scene text recognition on complicated and/or low-quality images, and propose a novel method FAN to solve this problem. Different from the existing methods, FAN uses an innovative focusing network to rectify the drifted attention of the AN model in handling complicated and low-quality images. Extensive experiments over several benchmarks show that the proposed method significantly outperforms existing methods.
In the future, we plan to extend the proposed idea to text detection and other related tasks.

\section*{Acknowledgement}Fan Bai and Shuigeng Zhou were partially supported by the Key Projects of
Fundamental Research Program of Shanghai Municipal Commission of Science and Technology under grant No.~14JC1400300.

{\small
\bibliographystyle{ieee}
\bibliography{egbib}
}

\end{document}